# CF-NODE: Counterfactual Spatial Node Intervention for Test-Time Adaptation Under Domain Shift, Evaluated Across Five Diabetic Retinopathy Datasets

Sabarish G and Deepak Parashar, Senior Member, IEEE

***Abstract*—Automated fundus-based diabetic retinopathy (DR) grading can substantially expand screening capacity, but source-trained models degrade sharply across different cameras, clinics, and patient populations. Existing test-time adaptation methods depend on entropy minimization or target information accumulated from image batches and streams. These mechanisms can reinforce confidently incorrect predictions and break down when clinical images arrive independently. We introduce CF-NODE, a novel counterfactual test-time adaptation framework for cross-domain DR grading. Instead of trusting prediction confidence or estimating a target distribution, CF-NODE asks how each prediction responds under a controlled intervention on the evidence supporting it. CF-NODE ranks internal spatial nodes using model-derived attention, feature magnitude, and node-set similarity, progressively attenuates the highest-ranked nodes, and uses the stability of the resulting expected-grade response to apply a bounded calibration of the predicted grade distribution. The framework is fully frozen and adapts every image independently using only geometric views and a fixed source-derived prior, requiring no target-derived information or model updates. We evaluate CF-NODE on five external DR datasets under repeated-seed evaluation against TENT, EATA, SAR, and T3A. CF-NODE outperforms all four methods with statistically significant improvements over the second-best method, improves AUC on every target domain, and attains the highest average AUC and macro-F1, delivering stronger grade discrimination together with more balanced recognition across DR severity levels. These results demonstrate that counterfactual intervention sensitivity provides consistent and deployment-ready domain robustness for single-image clinical inference. Code is available at https://github.com/sabarishknight/CF-Node.**

## I. INTRODUCTION

DIABETIC Retinopathy (DR) is one of the major causes of preventable vision loss, and accurate grading of retinal fundus images is important for timely referral and treatment [1], [2]. Deep neural networks have achieved strong performance in automated DR screening and severity grading [3], [4], [5]. However, their reliability often deteriorates when they are applied outside the training domain. Differences in retinal cameras, illumination, field of view, acquisition protocols, image quality, and patient populations can alter the appearance of fundus images and lead to confident but incorrect grades [6], [7]. This limits the safe use of a model across clinics and screening programs, even when it performs well on its original test set.

Clinical deployment also creates a difficult inference setting. Fundus images may arrive individually, on demand, and from different cameras or clinical sites. A model may not have access to a homogeneous batch of target images, and it may be unrealistic to assume that consecutive patients belong to the same target distribution. Entropy-based TTA methods [8], [9], [10] can also be risky in this setting. Low entropy indicates confidence rather than correctness, so entropy minimization may reinforce a confident but incorrect grade under domain shift. We therefore consider the following problem: how can a source-trained DR grader improve its prediction for one unlabeled image from an unseen domain without entropy minimization, parameter updates, target labels, target-domain statistics, a target batch, or an adaptation stream?

Test-time adaptation (TTA) attempts to improve model performance using unlabeled images observed during inference. A common approach is entropy minimization. TENT adapts batch-normalization parameters by reducing prediction entropy [8], while EATA combines reliable-sample filtering with anti-forgetting regularization, whereas SAR uses reliable-sample filtering and sharpness-aware updates [9], [10]. These methods can be effective, but entropy is not a direct measure of prediction correctness. A shifted model may assign high confidence to an incorrect grade and therefore produce low entropy. In this case, entropy provides little warning that correction is needed and minimizing it can make the incorrect prediction even more confident [11]. Online parameter updates may also allow an unreliable image or an unstable target batch to influence predictions for later patients.

Another group of TTA methods uses target features, class supports, prototypes, memory banks, or estimated target statistics. T3A updates class supports using incoming target features [12], while FunOTTA maintains a target-feature memory bank and uses prototype-based adaptation for fundus images [11]. Such methods can benefit from repeated observations from a stable target domain. However, their behavior can depend on pseudo-label quality, batch composition, stream order, and previously observed images [11], [12]. These assumptions may not hold when clinical images arrive independently or originate from different devices and sites. Estimating a target-class distribution or target feature structure from a single image is also unreliable. These constraints motivate a single-image method that does not rely on target-derived statistics or adaptation state carried across patients.

We introduce Counterfactual Spatial Node Intervention (CF-NODE), a frozen-model approach that obtains an adaptation signal directly from the current image. Instead of updating model parameters from entropy or estimating a target distribution, CF-NODE asks a different question: how does the predicted DR grade change when the spatial representation selected by the model is deliberately weakened? The final convolutional feature map is represented as a set of spatial nodes. CF-NODE ranks these nodes using learned attention, feature magnitude, and similarity to the complete node set, then progressively attenuates the highest-ranked nodes across four intervention levels. The magnitude and consistency of the expected-grade decrease across intervention levels define a per-image counterfactual sensitivity score. This score measures dependence within the trained model.

CF-NODE uses intervention sensitivity to control bounded, image-specific probability calibration. A small intervention response keeps the calibrated distribution close to the source prediction, while a stronger and more stable response produces a larger but bounded change in its concentration. Geometric views of the same image are then aggregated, and a fixed source-derived class prior is applied without estimating statistics from the target dataset. The complete model remains frozen, and each image is processed independently without target labels, target batches, gradients, optimization, or information from previous patients. This avoids cross-patient state and prevents an unreliable test image from changing the model used for subsequent cases.

The key distinction from existing TTA methods is the source of the adaptation signal. Entropy-based methods adapt from prediction confidence, while prototype and memory methods adapt from accumulated target features. CF-NODE instead creates a controlled intervention inside the spatial representation and uses the measured response to control the strength of bounded probability calibration. The main contributions are:

- We introduce CF-NODE, a novel counterfactual test-time adaptation method for cross-domain DR grading. It refines each image-specific prediction through interventions on ranked internal spatial nodes.
- We develop a model-guided framework that first ranks internal spatial nodes using attention, feature magnitude, and global node-set similarity, then applies graded counterfactual interventions to the highest-ranked nodes and converts the stability-normalized expected-grade response into a bounded probability correction.
- CF-NODE enables fully frozen, independent-image deployment. Each prediction uses only the frozen source model, geometric views of the current image, and a fixed source-derived prior, requiring no target labels, cross-image target batches or statistics, memory banks, pseudo-labels, gradients, model updates, or persistent adaptation state.
- Across five external DR datasets, CF-NODE improves AUC on every target and achieves the highest mean AUC and macro-F1 among the evaluated methods. Relative to source-only inference, it gains 2.1 and 6.2 percentage points on average, demonstrating stronger grade discrimination and class-balanced recognition under domain shift.

## II. Related Work

CF-NODE is related to four areas of prior work: deep learning for DR grading, domain shift and test-time adaptation, counterfactual intervention, and source-prior adjustment.

Deep learning has been widely studied for automated DR screening and grading. Gulshan et al. [3] and Ting et al. [4] demonstrated strong diagnostic performance across large and diverse fundus image collections, while Krause et al. [13] highlighted the importance of expert reference-label quality. Zoom-in networks [14] and class-attention modules [15] further improve grading by emphasizing lesion-related regions. However, these approaches mainly improve feature extraction and do not directly address unseen domain shifts during deployment.

DR models often degrade across cameras, illumination conditions, acquisition protocols, and patient populations [7], [16]. Domain adaptation and transfer learning can reduce these differences [17] but generally require target samples and an additional adaptation stage. CF-NODE considers the stricter setting in which target images arrive independently and the target distribution is unavailable.

Test-time adaptation modifies model behavior during inference without target labels. TENT [8] minimizes entropy by updating batch-normalization parameters, EATA [9] adds reliable-sample filtering and anti-forgetting regularization, and SAR [10] uses sharpness-aware adaptation. These methods rely on confidence-based optimization and parameter updates, through which an unreliable update may affect subsequent predictions. T3A [12], LAME [18], and FunOTTA [11] instead use target supports, batch relationships, prototypes, or memory banks. Their behavior can therefore depend on multiple target samples, stream order, or accumulated information. In contrast, CF-NODE derives its adaptation signal from the current image while keeping the model frozen, requiring no target batch, accumulated features, statistics, or cross-image state.

Counterfactual reasoning examines how an outcome changes under an intervention [19], [20], [21]. Attribution methods such as Grad-CAM [22] and integrated gradients [23], together with attention-based multiple-instance learning [24], identify image regions that influence a prediction but are mainly used for interpretation or localization. CF-NODE instead uses the response to controlled intervention on ranked internal spatial nodes to refine predictions under domain shift. This model-level intervention measures dependence within the trained network and does not identify clinical lesions or establish clinical causality. The recent FACTOR preprint [25] uses counterfactual input perturbations for open-vocabulary object detection, whereas CF-NODE performs representation-level intervention for image-level DR grading.

Class prevalence can also affect predictions. Logit adjustment incorporates class-frequency information into classifier outputs [26], but estimating a target prior requires multiple target samples and is unreliable for independent-image inference. Because the balanced EyePACS training subset removes the original source proportions, CF-NODE applies a weighted source-derived log-prior after multi-view aggregation. It uses only stored source frequencies and does not estimate or claim to recover the target-class prevalence.

## III. Method

Fig. 1 provides an overview of CF-NODE. Given an unlabeled image from an unseen domain, the frozen source model produces a class distribution and a set of spatial feature nodes. CF-NODE ranks these nodes and attenuates the highest-ranked subset at several intervention levels. The changes in the model output are summarized as a per-image intervention-sensitivity score. This score controls a sensitivity-dependent temperature that calibrates the class distribution. The calibrated distributions are then aggregated across geometric views and adjusted using a fixed source-derived class prior. No target labels, target-domain statistics, gradients, or parameter updates are used.

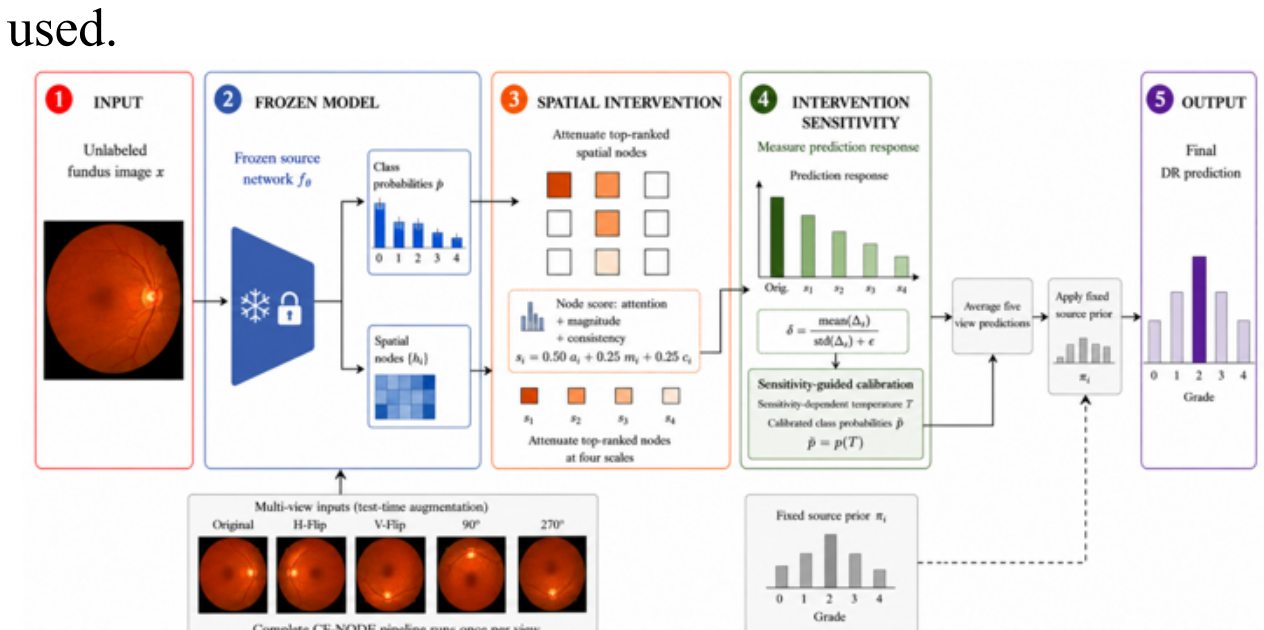

Fig. 1. Block diagram of the proposed CF-NODE framework.

### A. Source Model and Spatial Representation

CF-NODE requires a source model that exposes spatial feature vectors in addition to its class prediction. For an image $x \in \mathbb{R}^{3\times H\times W}$, the encoder produces a feature map $F = f_\theta^{\text{enc}}(x) \in \mathbb{R}^{C\times h\times w}$. Let $u_i \in \mathbb{R}^C$ denote the feature vector at the $i$-th spatial position, where $N = hw$. The spatial features are projected into $d$-dimensional node embeddings:

$$v_i = \text{GELU}\big(\text{LN}(W_n u_i + b_n)\big), \qquad i = 1, \dots, N \qquad (1)$$

Here, $W_n \in \mathbb{R}^{d\times C}$, $b_n \in \mathbb{R}^d$, and $v_i \in \mathbb{R}^d$. The projection parameters are learned during source training and frozen during test-time inference. The resulting node set is

$$V(x) = \{v_i\}_{i=1}^N \qquad (2)$$

Each $v_i$ represents the encoder features associated with one receptive-field region of the image. A node is an internal representation and is not assumed to correspond directly to a pathological structure.

A learned attention module assigns an unnormalized scalar score $e_i = q_\psi(v_i)$ to each node, where $q_\psi: \mathbb{R}^d \to \mathbb{R}$ is the attention-scoring module. The module is learned during source training and frozen during inference. The scores are normalized into attention weights:

$$a_i = \frac{\exp(e_i)}{\sum_{j=1}^N \exp(e_j)}, \qquad i = 1, \dots, N \qquad (3)$$

The attention-weighted nodes form a node-level descriptor $z_n \in \mathbb{R}^d$:

$$z_n = \sum_{i=1}^N a_i\, v_i \qquad (4)$$

A second branch obtains a global descriptor through generalized mean pooling and a learned global projection. The global descriptor $z_g \in \mathbb{R}^d$ is fused with the node-level descriptor:

$$z_g = g_\theta^{\text{glob}}\big(\text{GeM}(F)\big), \qquad z = f_\theta^{\text{fuse}}\big([z_n, z_g]\big) \qquad (5)$$

Here, $g_\theta^{\text{glob}}$ is the learned global projection, $[z_n, z_g]$ denotes feature concatenation, and $f_\theta^{\text{fuse}}$ is the learned fusion block. The GeM exponent is initialized to 3.0, learned during source training, and frozen during test-time inference.

The fused representation is passed to a $K$-class prediction head to obtain the class logits $\ell^{\text{cls}} \in \mathbb{R}^K$ and baseline class probabilities $p \in [0,1]^K$:

$$\ell^{\text{cls}} = g_\theta^{\text{cls}}(z), \quad p = \text{softmax}\left(\frac{\ell^{\text{cls}}}{T_0}\right), \quad T_0 = 1.05 \qquad (6)$$

Here, $T_0$ is the fixed default temperature selected using only the EyePACS validation split. All source-model components remain frozen during test-time inference. The source architecture and training procedure are described in Section IV. For the present DR application, $K = 5$.

### B. Counterfactual Spatial Node Intervention

**1) Node Ranking:** Before intervention, CF-NODE assigns each node a score using three model-derived signals: attention weight, feature magnitude, and similarity to the complete node set. The latter two are normalized within each image. The normalized feature magnitude is

$$m_i = \frac{\|v_i\|_2}{\max_j \|v_j\|_2 + \epsilon} \qquad (7)$$

The mean cosine similarity of node $v_i$ to the complete node set is

$$s_i = \frac{1}{N}\sum_{j=1}^N \frac{v_i^\top v_j}{\max(\|v_i\|_2, \epsilon_c)\max\left(\|v_j\|_2, \epsilon_c\right)} \qquad (8)$$

The similarity values are normalized within the image:

$$c_i = \frac{s_i - \min_j s_j}{\max_j s_j - \min_j s_j + \epsilon} \qquad (9)$$

The final node-importance score is

$$\phi_i = w_a a_i + w_m m_i + w_c c_i \qquad (10)$$

Here, $\epsilon = 10^{-8}$ and $\epsilon_c = 10^{-12}$ prevent division by zero. We use the fixed weights $w_a = 0.50$ and $w_m = w_c = 0.25$, selected using only the EyePACS validation split. The score $c_i$ measures similarity to all nodes in the image rather than only spatial neighbors. Nodes are ranked in descending order of $\phi_i$. This ranking selects the representations to be intervened on and does not determine whether they contain pathological or domain-specific information.

**2) Multi-Level Intervention:**

Let

$$\mathcal{M} = \{(r_m, \rho_m)\}_{m=1}^{M} \qquad (11)$$

denote $M$ intervention levels, where $0 < r_m \leq 1$ is the selected node fraction and $0 \leq \rho_m \leq 1$ is the retained magnitude. We use $M = 4$ with $(r_m, \rho_m) \in (0.02, 0.75), (0.05, 0.40), (0.10, 0.20), (0.15, 0.10)$. If $S_m$ contains the $k_m = \max(\lfloor r_m N \rfloor, 1)$ highest-ranked nodes, the intervened node representation is

$$v_i^{(m)} = \begin{cases} \rho_m v_i, & i \in S_m \\ v_i, & i \notin S_m \end{cases} \qquad (12)$$

The resulting node set is $V^{(m)} = {v_i^{(m)}}_{i=1}^{N}$. For each level, the attention pooling, fusion block, and classifier are reevaluated using $V^{(m)}$, while the global descriptor $z_g$ remains fixed. Attention weights are recomputed, allowing the frozen attention module to respond to the modified node set.

Let $p$ be the original class distribution in (6). After intervention level $m$, let $\ell^{\mathrm{cls},(m)}$ denote the reevaluated class logits. The corresponding distribution is

$$p^{(m)} = \mathrm{softmax}\left(\frac{\ell^{\mathrm{cls},(m)}}{T_0}\right) \qquad (13)$$

where $T_0 = 1.05$ is the same default temperature used for the original and intervened distributions. The expected grade is

$$h(p) = \sum_{k=0}^{K-1} k\, p_k \qquad (14)$$

The one-sided response at intervention level (m) is

$$\Delta_m = \left[h(p) - h\left(p^{(m)}\right)\right]_+ \qquad (15)$$

$$\mu_\Delta = \frac{1}{M}\sum_{m=1}^{M} \Delta_m \qquad (16)$$

$$\sigma_\Delta = \sqrt{\frac{1}{M-1}\sum_{m=1}^{M} (\Delta_m - \mu_\Delta)^2} \qquad (17)$$

The stable intervention-sensitivity score is

$$\delta = \frac{\mu_\Delta}{\sigma_\Delta + \epsilon_\Delta} \qquad (18)$$

where $\epsilon_\Delta = 0.15$ prevents instability when the responses have low variance. A high $\delta$ indicates that the mean one-sided reduction is large relative to its variation across intervention levels. Because the intervention modifies only selected nodes while the image, model parameters, global descriptor, and remaining nodes stay fixed, $\delta$ measures directional sensitivity to the selected representation. It does not establish lesion importance or clinical causality.

### C. Sensitivity-Dependent Calibration:

The stable intervention-sensitivity score $\delta$ determines a bounded temperature for the class distribution of the current view:

$$T_\delta = \mathrm{clip}(T_0 - 0.22\delta, 0.72, 1.08) \qquad (19)$$

The calibrated per-view distribution is

$$p_\theta^{\mathrm{view}} = \mathrm{softmax}\left(\frac{\log(p + \epsilon)}{T_\delta}\right) \qquad (20)$$

Here, $\mathrm{clip}(u, l, h) = \min(\max(u, l), h)$ and $\epsilon = 10^{-8}$. When the intervention response is small, $T_\delta$ remains close to the default temperature $T_0$. A larger stable response lowers $T_\delta$ and increases the concentration of the class distribution within fixed bounds. All constants are selected using only the EyePACS validation split. The procedure uses only the current image and frozen source model, without target labels, target-domain statistics, gradients, parameter updates, or cross-image adaptation state. The resulting $p_\theta^{\mathrm{view}}$ is passed to multi-view aggregation.

### D. Multi-View Inference and Source-Prior Adjustment

The complete procedure is applied independently to a set $\mathcal{A}$ of label-preserving geometric views of the same image. For each transformation $\tau \in \mathcal{A}$, $\tau(x)$ denotes the corresponding view of image $x$. The calibrated distributions are averaged:

$$p^{\mathrm{mv}} = \frac{1}{|\mathcal{A}|}\sum_{\tau \in \mathcal{A}} p_\theta^{\mathrm{view}}\left(\tau(x)\right) \qquad (21)$$

Here, $|\mathcal{A}|$ is the number of geometric views. A fixed source-prior adjustment is then applied:

$$p^{\mathrm{corr}} = \mathrm{softmax}(\log(p^{\mathrm{mv}} + \epsilon) + \eta \log(\pi_s + \epsilon)) \qquad (22)$$

Here, $\epsilon = 10^{-8}$, and $\pi_{s,k} = n_{s,k} / \sum_{j=0}^{K-1} n_{s,j}$, where $n_{s,k}$ is the number of Grade (k) images in the full EyePACS training set before balanced subsampling. We use $\pi_s = (0.734783, 0.069550, 0.150658, 0.024853, 0.020156)$ and $\eta = 0.18$. Both logarithms are applied elementwise. The source prior is fixed before external evaluation and is not estimated from target images. The final grade prediction is

$$\hat{y} = \underset{k \in \{0,\ldots,K-1\}}{\mathrm{argmax}}\; p_k^{\mathrm{corr}} \qquad (23)$$

For the present DR application, CF-NODE uses five geometric views: the original image, horizontal flip, vertical flip, 90° rotation, and 270° rotation. Each view is processed independently through the frozen model and intervention procedure before probability averaging.

For computational efficiency, the four intervention sets for each view are concatenated and evaluated together through the attention, fusion, and classification blocks. All operations remain sample-wise, so this temporary batching introduces no target-domain statistics or cross-image adaptation state. The source model remains in evaluation mode throughout inference.

## IV. Experiment Setup

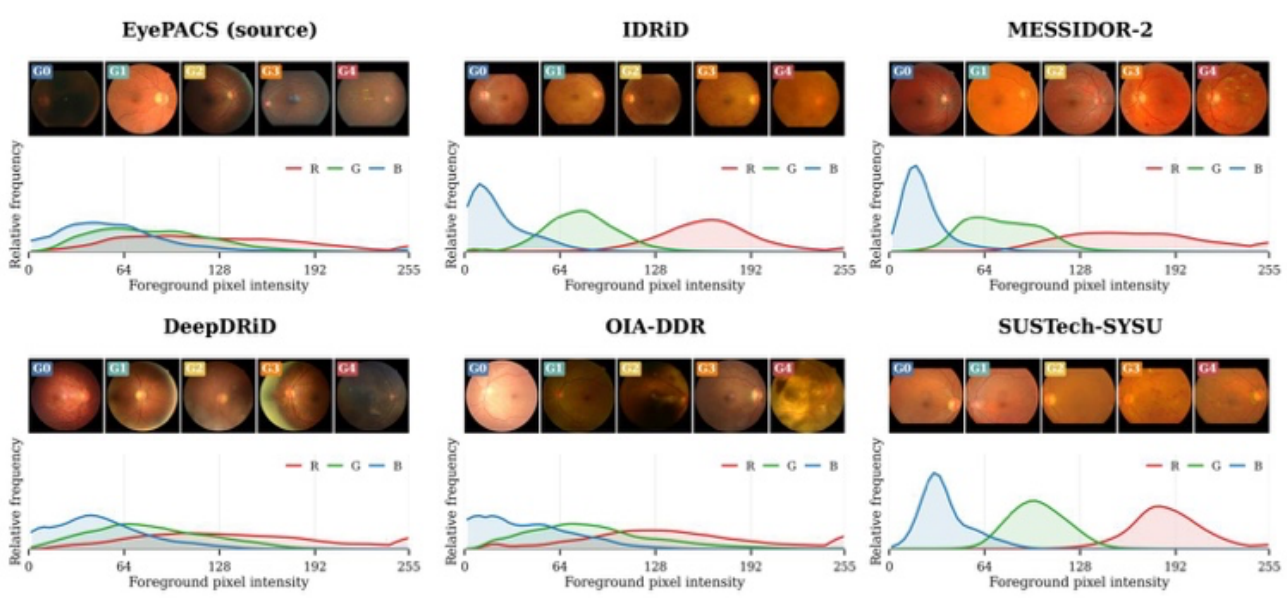

Fig. 2. Fundus images and RGB foreground pixel-intensity distributions across the source and target datasets

We test CF-NODE in a cross-domain setting. A source model is trained on EyePACS and evaluated on five external DR datasets collected using different acquisition systems. No target-domain image or label is used during source training, hyperparameter selection, or test-time correction.

### A. Datasets

The source domain is the labeled EyePACS training set [27], which contains 35,126 fundus images graded from 0 to 4. Following the source-training protocol used by FunOTTA [11], we select a balanced subset of 3,540 images, with 708 images per grade. This subset is divided into 2,832 training and 708 validation images using an 80/20 split. Cross-domain evaluation is conducted on five target datasets: IDRiD [28], containing 516 images; MESSIDOR-2 [29] containing 1,748 images, of which 1,744 gradable images are used; OIA-DDR [30], containing 13,673 images, of which 12,522 images labeled from Grade 0 to Grade 4 are used; DeepDRiD [6], containing 2,000 images, from which the official validation set of 400 images is used; and SUSTech-SYSU [31], containing 1,219 images, of which 1,151 images labeled from Grade 0 to Grade 4 are used after excluding 68 images with laser spots or scars. Representative images across Grades 0 to 4 and the foreground RGB intensity distributions of the source and target datasets are shown in Fig. 2. These datasets cover different clinical sites and image-acquisition systems. Every target image is processed independently, without target batches or previously accumulated target information. Target labels are not used during adaptation and are accessed only after inference to calculate the evaluation metrics.

### B. Evaluation Metrics

The target datasets have different class distributions, with substantially fewer examples of severe DR grades. We therefore report macro-averaged one-vs-rest AUC and macro-F1. AUC is calculated from the normalized five-class probabilities, with classes absent from a target dataset excluded from its macro average. Macro-F1 is calculated from the discrete argmax predictions. Results are reported for each target dataset and as the mean across all five datasets. All values are presented as mean ($\pm$) standard deviation over five random seeds.

### C. Implementation Details

The source model uses an ImageNet-initialized ResNet-50 backbone [32] with $320 \times 320$ inputs. Its final $10 \times 10 \times 2048$ feature map provides 100 spatial nodes, each projected to a 128-dimensional embedding. The nodes are aggregated using an attention-based DeepSets module [33], while a learnable generalized mean-pooling layer [34] produces a global descriptor. The node and global descriptors are fused and passed to a five-class classification head.

The training objective combines cross-entropy with label smoothing $\epsilon = 0.05$ and focal loss with $\gamma = 1.8$. Only the task-specific layers above the backbone are trained during the first four epochs, followed by the fourth ResNet stage from epochs 5 to 14 and the complete network from epoch 15 onward. We use AdamW, cosine-annealing warm restarts, weight decay of $8 \times 10^{-5}$, gradient clipping at 1.0, and an exponential moving average with decay 0.999. Training uses a batch size of 16 for at most 40 epochs, with early stopping on the EyePACS validation set using a patience of 15 epochs. The best exponential-moving-average checkpoint is used for all target-domain experiments.

Training augmentation consists of random resized cropping, horizontal flipping, mild rotation, light color jitter, and occasional Gaussian blur. Validation and base test images are resized to $320 \times 320$ and normalized using ImageNet statistics. Experiments are implemented in PyTorch and run using two NVIDIA Tesla T4 GPUs.

### D. Baselines

We compare CF-NODE with the source-only model, TENT [8], EATA [9], SAR [10], and T3A [12]. For each training seed, all methods use the same source checkpoint, input preprocessing, target-image order, and evaluation code. Method-specific hyperparameters are selected using only the EyePACS validation split and fixed before external evaluation. A fresh copy of the corresponding source checkpoint is loaded for every method, target dataset, and training-seed run. Target images are processed in their original order without shuffling. For methods that maintain an online adaptation state, the state is reset at the beginning of each target dataset.

In our study, TENT, EATA, and SAR are slightly modified because of the high sensitivity of the DR grading model to parameter changes. When reproduced with their original update rules, these methods show unstable and degraded performance in our setting. By updating only the classifier weights, we obtain a clear improvement, and we therefore report this stronger configuration, marked by † in Tables II and III. The same reproduction strategy is used for these baselines in FunOTTA [11].

For CF-NODE, the source model remains in evaluation mode, and inference is performed without gradients or parameter updates. Five geometric views are generated for each image: the original image, horizontal flip, vertical flip, 90° rotation, and 270° rotation. Each view is processed independently through the complete CF-NODE pipeline.

Node importance is computed as a weighted combination of attention, feature magnitude, and representation consistency, with weights of 0.50, 0.25, and 0.25, respectively. We use four intervention configurations: 0.02,0.75, 0.05,0.40, 0.10,0.20, and 0.15,0.10. In each pair, the first value specifies the proportion of top-ranked nodes selected for intervention, and the second specifies their multiplicative attenuation factor. All CF-NODE settings, including the intervention configurations, node-ranking weights, temperature parameters, and source-prior strength, are selected using only the EyePACS validation split. The intervention-sensitivity score determines the sensitivity-dependent temperature according to the calibration rule defined in Section III.

The calibrated class distributions from the five geometric views are averaged. We then apply a fixed source-prior adjustment with strength 0.18 using the stored source-derived prior.

## V. Results and Discussion

Tables II and III compare CF-NODE with the source-only ResNet-50 and four TTA methods, TENT [8], EATA [9], SAR [10], and T3A [12], across IDRiD, MESSIDOR-2, OIA-DDR, DeepDRiD, and SUSTech-SYSU. All methods use the same source checkpoints, preprocessing, target-image order, and evaluation pipeline. Method-specific hyperparameters are selected using only the EyePACS validation split and are applied unchanged to all five target domains.

CF-NODE achieves the highest average AUC of 80.3% and macro-F1 of 42.1%. Compared with the source-only model, this corresponds to improvements of 2.1 percentage points in AUC and 6.2 percentage points in macro-F1. Its AUC gains are 2.9, 1.2, 1.6, 2.1, and 2.4 percentage points on IDRiD, MESSIDOR-2, OIA-DDR, DeepDRiD, and SUSTech-SYSU, respectively. CF-NODE obtains the highest AUC on all five datasets and the highest macro-F1 on IDRiD, MESSIDOR-2, and OIA-DDR. Its average improvement over the second-best method is statistically significant for both AUC $p \leq 0.05$ and macro-F1 $p \leq 0.01$.

The baseline methods show different behavior across the two metrics. TENT maintains the same average AUC as the source model at 78.2%, while improving macro-F1 from 35.9% to 38.7%. EATA produces a similar result, with an average AUC of 78.1% and macro-F1 of 37.4%. SAR improves macro-F1 to 37.5% but reduces average AUC to 77.6%, while T3A obtains the second-highest average AUC of 79.1% but a macro-F1 of 36.1%. Because AUC measures class-probability ranking and macro-F1 measures discrete grade assignments, improvements in one metric do not necessarily produce comparable improvements in the other. Fig. 3 shows these per-domain differences, where the mean ± SD interval of CF-NODE lies entirely above zero for AUC, while TENT, EATA, and SAR span both sides of zero.

CF-NODE provides the strongest average performance across both metrics. The simultaneous improvement in AUC and macro-F1 is important for cross-domain DR grading because AUC measures ranking performance, while macro-F1 gives equal importance to each grade and reduces the influence of the dominant Grade 0 class. On DeepDRiD, CF-NODE improves macro-F1 over the source model from 37.9% to 39.5% and achieves the highest AUC of 77.0%, although TENT obtains a higher macro-F1 of 41.2%. On SUSTech-SYSU, CF-NODE achieves the highest AUC of 82.4% and the highest macro-F1 among the adaptation methods at 37.3% but remains 1.1 percentage points below the source model's macro-F1 of 38.4%. This result is consistent with the strongly imbalanced grade distribution of SUSTech-SYSU [31], which makes macro-F1 sensitive to a small number of errors in minority grades. As Fig. 3 shows, every evaluated method loses macro-F1 on this domain, and CF-NODE loses the least. Since CF-NODE does not estimate target-specific class frequencies, improved probability ranking may not always translate into the target-specific grade boundaries that maximize macro-F1. Nevertheless, its highest AUC on every target dataset and its highest average performance across both metrics support more consistent domain robustness, while the DeepDRiD and SUSTech-SYSU results show that dataset-specific discrete grade decisions remain challenging.

To isolate the contribution of counterfactual node selection, we keep the geometric views, fixed source prior, source checkpoints, and evaluation protocol unchanged. Tables IV and V compare the fixed scaffold without node intervention, random-node intervention, and full CF-NODE with ranked node selection. Results are averaged over five seeds across all five target domains. All variants operate without target labels, target-domain statistics, gradients, or parameter updates. The fixed scaffold achieves 79.7% average AUC and 40.7% macro-F1, while random-node intervention gives nearly identical results of 79.8% and 40.7%, showing that arbitrary intervention provides little benefit. Full CF-NODE improves these averages to 80.3% and 42.1%, gains of 0.6 AUC and 1.4 macro-F1 points. It improves AUC on all five domains and macro-F1 on four, with the largest macro-F1 gains on MESSIDOR-2 and IDRiD. Although random selection gives marginally higher AUC on DeepDRiD and SUSTech-SYSU, and SUSTech-SYSU macro-F1 remains nearly unchanged, these differences do not improve average performance. Overall, the results confirm that the gains come from ranked counterfactual node selection rather than arbitrary node attenuation.

## VI. Conclusion

In this paper, we introduced CF-NODE, a counterfactual node-based test-time adaptation framework for robust diabetic retinopathy grading under domain shift. Unlike methods that rely mainly on prediction confidence or accumulated target statistics, CF-NODE examines how predictions respond when ranked disease-related spatial nodes are progressively attenuated. This counterfactual evidence is combined with a fixed source prior to adapt each target image independently, without target labels, target-domain statistics, gradients, parameter updates, or cross-patient adaptation state.

We validated CF-NODE across five diverse fundus image datasets: IDRiD, MESSIDOR-2, OIA-DDR, DeepDRiD, and SUSTech-SYSU. CF-NODE achieved the best average performance among all evaluated methods, reaching 80.3% AUC and 42.1% macro-F1, with improvements of 2.1 and 6.2 percentage points over the source-only model. It obtained the highest AUC on all five target domains and the highest macro-F1 on three, with statistically significant improvements over the second-best method. The ablation results further confirm that the gains come from ranked counterfactual node selection rather than arbitrary node intervention. Overall, these findings demonstrate the effectiveness and consistency of CF-NODE for single-image adaptation across different acquisition settings.

The current evaluation focuses on closed-set diabetic retinopathy grading using fundus images. Future work will investigate the applicability of CF-NODE to other retinal conditions, such as glaucoma and age-related macular degeneration, and to broader medical imaging tasks under domain shift. We hope that this work supports further development of reliable counterfactual test-time adaptation methods for medical image analysis.

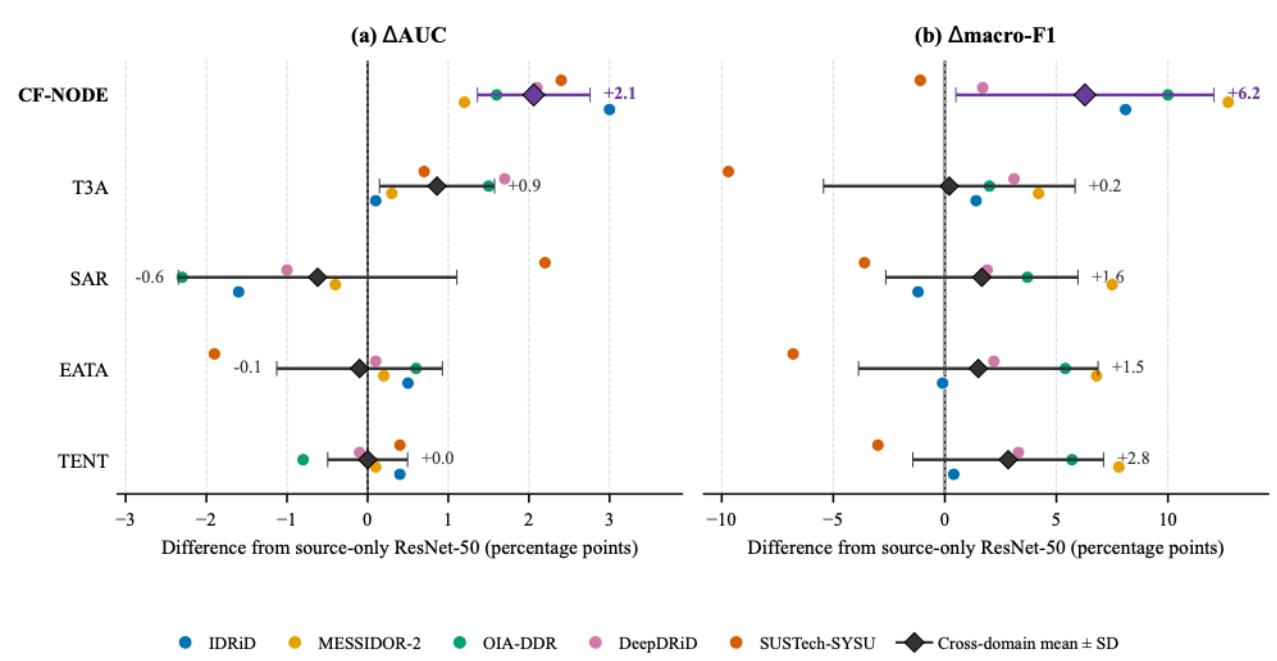

Fig. 3. Domain-wise AUC and macro-F1 differences from source-only ResNet-50. Diamonds and whiskers show the cross-domain mean ±SD; positive values favor adaptation.

TABLE II
AUC COMPARISON ACROSS FIVE TARGET DOMAINS

| Dataset | IDRiD | MESSIDOR -2 | OIA-DDR | DeepDRiD | SUSTech -SYSU | Avg |
|---|---|---|---|---|---|---|
| Metric | AUC | AUC | AUC | AUC | AUC | AUC |
| ResNet50 | $80.8_{1.6}$ | $78.7_{0.9}$ | $76.7_{1.0}$ | $74.9_{1.3}$ | $80.0_{2.1}$ | 78.2 |
| TENT† [8] | $81.2_{1.8}$ | $78.8_{1.3}$ | $75.9_{1.0}$ | $74.8_{1.1}$ | $80.4_{2.5}$ | 78.2 |
| EATA† [9] | $\underline{81.3}_{1.7}$ | $78.9_{1.0}$ | $77.3_{1.2}$ | $75.0_{1.7}$ | $78.1_{2.2}$ | 78.1 |
| SAR† [10] | $79.2_{1.9}$ | $78.3_{1.2}$ | $74.4_{1.2}$ | $73.9_{1.3}$ | $\underline{82.2}_{1.9}$ | 77.6 |
| T3A [12] | $80.9_{2.7}$ | $\underline{79.0}_{0.5}$ | $\underline{78.2}_{1.0}$ | $\underline{76.6}_{1.1}$ | $80.7_{1.3}$ | $\underline{79.1}$ |
| **CF-NODE (ours)** | $\mathbf{83.7}_{2.1}$ | $\mathbf{79.9}_{1.1}$ | $\mathbf{78.3}_{1.2}$ | $\mathbf{77.0}_{1.4}$ | $\mathbf{82.4}_{3.7}$ | $\mathbf{80.3}^{*}$ |

TABLE III
MACRO-F1 COMPARISON ACROSS FIVE TARGET DOMAINS

| Dataset | IDRiD | MESSIDOR -2 | OIA-DDR | DeepDRiD | SUSTech -SYSU | Avg |
|---|---|---|---|---|---|---|
| Metric | F1 | F1 | F1 | F1 | F1 | F1 |
| ResNet50 | $40.0_{1.5}$ | $34.1_{1.5}$ | $28.9_{1.0}$ | $37.9_{0.8}$ | $\mathbf{38.4}_{1.4}$ | 35.9 |
| TENT† [8] | $40.4_{2.1}$ | $\underline{41.9}_{3.2}$ | $\underline{34.6}_{1.8}$ | $\mathbf{41.2}_{2.1}$ | $35.4_{1.6}$ | $\underline{38.7}$ |
| EATA† [9] | $39.9_{2.4}$ | $40.9_{3.2}$ | $34.3_{2.3}$ | $40.1_{3.0}$ | $31.6_{1.8}$ | 37.4 |
| SAR† [10] | $38.8_{2.8}$ | $41.6_{3.4}$ | $32.6_{2.1}$ | $39.8_{1.8}$ | $34.8_{1.4}$ | 37.5 |
| T3A [12] | $\underline{41.4}_{2.3}$ | $38.3_{2.5}$ | $30.9_{2.4}$ | $\underline{41.0}_{2.7}$ | $28.7_{2.1}$ | 36.1 |
| **CF-NODE (ours)** | $\mathbf{48.0}_{4.3}$ | $\mathbf{46.8}_{2.9}$ | $\mathbf{38.8}_{2.2}$ | $39.5_{2.2}$ | $\underline{37.3}_{3.1}$ | $\mathbf{42.1}^{**}$ |

Note (table II & III): Values are reported as $\text{mean}_{\text{SD}}$ over five seeds, with SD shown as a subscript. Bold and underlined means indicate the best and second-best results, respectively. The dagger symbol (†) indicates that the method was adapted to our setting by fine-tuning only the classifier weights. * and ** indicate significance against the second-best method using a paired t-test on matched seed-level results ($p \leq 0.05$) and ($p \leq 0.01$).

TABLE IV
AUC ABLATION OF COUNTERFACTUAL NODE SELECTION ACROSS TARGET DOMAINS

| Configuration | Node selection | MESSIDOR-2 | IDRiD | DeepDRiD | OIA-DDR | SUSTech-SYSU | Avg. |
|---|---|---|---|---|---|---|---|
| Metric | | AUC | AUC | AUC | AUC | AUC | AUC |
| Fixed scaffold | None | 79.24 | 82.35 | 76.79 | 77.50 | 82.37 | 79.7 |
| Random-node CF | Random | 79.23 | 82.40 | **77.14** | 77.57 | **82.48** | 79.8 |
| Full CF-NODE | Ranked | **79.91** | **83.75** | 77.01 | **78.31** | 82.41 | **80.3** |

TABLE V
MACRO-F1 ABLATION OF COUNTERFACTUAL NODE SELECTION ACROSS TARGET DOMAINS

| Configuration | Node selection | MESSIDOR-2 | IDRiD | DeepDRiD | OIA-DDR | SUSTech-SYSU | Avg. |
|---|---|---|---|---|---|---|---|
| Metric | | F1 | F1 | F1 | F1 | F1 | F1 |
| Fixed scaffold | None | 44.18 | 46.09 | 38.07 | 37.99 | **37.36** | 40.7 |
| Random-node CF | Random | 44.18 | 46.03 | 38.23 | 37.81 | 37.24 | 40.7 |
| Full CF-NODE | Ranked | **46.83** | **48.08** | **39.55** | **38.85** | 37.35 | **42.1** |